\documentclass[preprint,12pt,authoryear]{elsarticle}

\usepackage{amssymb}

\usepackage{easy-todo}
\usepackage{latexsym}
\usepackage{amsmath}
\usepackage{amsfonts}
\usepackage{algorithmic}
\usepackage{algorithm}
\usepackage{amssymb}
\usepackage{amsthm}
\usepackage{epsfig}
\graphicspath{ {.} }
\usepackage{graphics}
\usepackage{graphicx}
\usepackage{caption}
\usepackage{subcaption}
\usepackage{color}
\usepackage{float}
\usepackage[usenames,dvipsnames]{xcolor}
\usepackage{bm}

\usepackage{gensymb}

\newcommand{\bC}{\bm{C}}

\newcommand{\bK}{\bm{K}}
\newcommand{\bk}{\bm{k}}

\newcommand{\bH}{\bm{H}}

\newcommand{\bv}{\bm{v}}

\newcommand{\bx}{\bm{x}}
\newcommand{\bX}{\bm{X}}
\newcommand{\by}{\bm{y}}
\newcommand{\bY}{\bm{Y}}
\newcommand{\bz}{\bm{z}}

\newcommand{\bP}{\bm{P}}
\newcommand{\bT}{\bm{T}}

\newcommand\floor[1]{\lfloor#1\rfloor}
\newcommand{\bSigma}{\bm{\Sigma}}

\newcommand{\bmu}{\bm{\mu}}

\newcommand{\Var}{\mbox{Var}}
\newcommand{\E}{\mbox{E}}
\usepackage{tikz}

\journal{Environmental Modelling \& Software}

\begin{document}

\begin{frontmatter}



\title{Environmental Modeling Framework using Stacked Gaussian Processes}

\author[label1]{Kareem Abdelfatah}
\author[label2]{Junshu Bao}
\author[label1]{Gabriel Terejanu}

\address[label1]{Dept. of Computer Science \& Engineering at University of South Carolina \\ krabea@email.sc.edu,terejanu@cec.sc.edu}
\address[label2]{Dept. of Mathematics and Computer Science at Duquesne University \\ baoj@duq.edu}

\begin{abstract}
A network of independently trained Gaussian processes (StackedGP) is introduced to obtain predictions of quantities of interest with quantified uncertainties. The main applications of the StackedGP framework are to integrate different datasets through model composition, enhance predictions of quantities of interest through a cascade of intermediate predictions, and to propagate uncertainties through emulated dynamical systems driven by uncertain forcing variables. By using analytical first and second-order moments of a Gaussian process with uncertain inputs using squared exponential and polynomial kernels, approximated expectations of quantities of interests that require an arbitrary composition of functions can be obtained. The StackedGP model is extended to any number of layers and nodes per layer, and it provides flexibility in kernel selection for the input nodes. The proposed nonparametric stacked model is validated using synthetic datasets, and its performance in model composition and cascading predictions is measured in two applications using real data. 
\end{abstract}


\end{frontmatter}


\section{Introduction}

Gaussian processes (GP) (\citet{williams1996gaussian,Rasmussen:1997:EGP:927743,rasmussen2006gaussian}) are nonparametric statistical models that compactly describe distributions over functions with continuous domains. They have found various applications in the environmental modeling community, where they are used as data-driven models capable to predict various quantities of interest with quantified uncertainties such as ultra fine particles (\citet{reggente2014prediction}), mean temperatures over North Atlantic Ocean (\citet{Higdon1998}), wind speed (\citet{Hu20151456}), and monthly streamflow (\citet{Sun201472}), just to name a few. When the training data for GPs comes from simulators rather than field measurements, then GPs become computational efficient surrogate models or emulators of high-fidelity models (\citet{Kennedy2002,OHAGAN20061290,Conti2010640}), with various applications in environmental modeling such as fire emissions (\citet{Katurji2015254}), ocean and climate circulation (\citet{doi:10.1175/JTECH-D-11-00110.1}), urban drainage (\citet{Machac201654}), and computational fluid dynamics (\citet{Moonen201577}).

This paper develops a general probabilistic modeling framework based on a network of independently trained GPs (StackedGP), see Fig.~\ref{fig:StackedGP}, to obtain approximated expectations of quantities of interest that require model composition. Information integration through model composition is common in geostatistics and environmental sciences. For example, many environmental models are obtained using a composition of phenomenological/physical models determined using wet-lab measurements and forcing models determined using geospatial observations (\citet{Letcher2009,Jorgensen2010}). Phenomenological/physical models describe relationships between forcing variables (e.g. temperature) and quantities of interest (e.g. accumulation of carcinogenic toxins in corn, \citet{LiP_SIGS_2015}). Forcing models are used to calculate forcing variables at a location of interest using spatial interpolations. The composition of the two type of models yields geospatial estimates for the quantities of interest. The central challenge is that there is a compound effect of uncertainties coming from interpolation errors and model errors that need to be quantified and exposed to the quantities of interest. Furthermore, this model composition can be arbitrary and highly nested to capture the phenomenon of interest and make prediction for potentially unobserved quantities of interest.

The proposed general predictive modeling framework, StackedGP, extends and unifies the work of \citet{girard2002gaussian} and \citet{LiP_SIGS_2015}. In \citet{LiP_SIGS_2015}, the authors introduced StackedGP to predict carcinogenic toxin concentrations using environmental conditions. Monte Carlo sampling was used to propagate the uncertainty through the stacked model and estimate the mean and variance of the quantity of interest. Since sampling requires a high computational cost, here, the uncertainty propagation through the network is achieved approximately by leveraging the exact moments for the predictive mean and variance derived by \citet{girard2002gaussian} for a single GP with uncertain inputs and squared exponential kernel. We provide a re-derivation of the expectations in the squared exponential kernel case and a novel derivation for the predictive mean and variance corresponding to the polynomial kernels. We emphasize the impact of input uncertainty on the predictive mean and variance, which is key in obtaining better predictions. Namely, the input uncertainty weighs the contributions of the particular input to the GP node's prediction. Finally, we extend the StackedGP model to any number of layers and nodes per layer and provide an algorithm to obtain approximated expectations of quantities of interest that require arbitrary composition of models. The StackedGP model is validated in the numerical results section using various synthetic datasets, and it is applied to estimate the burned area using meteorological data (\citet{cortez2007data, taylor2006science}).

StackedGP is conceptually different from deep GPs (\citet{damianou2013deep}), where no data is available for the latent nodes and where the latent variable model requires to jointly infer the hyperparameters corresponding to the mappings between the layers. A model carrying the same name was introduced by \citet{neumann2009stacked}, where a stacked Gaussian process was proposed to model pedestrian and public transit flows in urban areas. The model proposed by \citet{neumann2009stacked} is capable of capturing shared common causes using a joint Bayesian inference for multiple tasks. In our work, the inference is performed independently for each GP node and the uncertainty is approximately propagated through the network. StackedGP provides flexibility in kernel selection for intermediate nodes (RBF, polynomial as well as kernels obtained via their sum) and has no restriction in selecting a suitable kernel for input nodes. Since the GP nodes are independently trained using different datasets, the running time of the StackedGP grows linearly with the number of nodes and can be sped up through embarrassing parallel training of GPs. 

In addition to information integration through model composition, StackedGP can be used to enhance predictions of quantities of interest using intermediate predictions of auxiliary variables. The unobserved target variables in supervised learning problems are often split in primary/main variables or quantities of interest and secondary/auxiliary variables. Given that they are unobserved at testing inputs, the secondary variables are often discarded in the learning problem where the mapping between observed inputs and primary variables is inferred. Enhanced predictions of quantities of interest can be obtained by stacking GPs for predicting intermediate secondary responses that govern the input space of GPs used to predict primary responses. Several examples can illustrate the idea such as uranium spill accident (\citet{seeger2005}) and predicting cadmium concentration in Swiss Jura (\citet{goovaerts1997,wilson12icml}). \citet{wilson12icml} developed a Gaussian Process Regression Network to model the correlations between multiple outputs such as primary and secondary responses. The outputs are given by weighted linearly combinations of latent functions where GP priors are defined over the weights, unlike  similar studies (\citet{seeger2005,boyle2005dependent}) where the weights are considered fixed. StackedGP is not designed to capture the correlations of response variables, however, StackedGP models can be constructed by stacking GPs for predicting intermediate secondary responses that govern the input space of GPs used to predict primary responses. This hierarchical framework outperforms other methods as described in the numerical results section, where Jura dataset (\citet{wilson12icml}) is used to assess the prediction accuracy of model with intermediate predictions.

Gaussian processes with uncertain inputs have been previously used in multi-step time series predictions (\citet{girard2003gaussian, candela2003propagation}). Modeling multi-step ahead predictions can be achieved by feeding back the predicted mean and variance at each time and propagating the uncertainty to the next time step. This idea has been used in different time-series applications such as electricity forecasting (\citet{lourencco2010short}) and water demand forecasting (\citet{wang2014gaussian}). In environmental sciences, uncertainty propagation through dynamical systems is also relevant when high-fidelity models are emulated (\cite{Castelletti20125, 10.2307/25464566, Conti2010640, bhattacharya2007}). For example, propagating uncertainties through atmospheric dispersion models (\cite{Nielsen1999, Sykes2006}) can be tackled through emulation. In this case, emulators can be used to speed up the uncertainty propagation process and obtain estimates of  quantities of interest with quantified uncertainties (\citet{konda2010uncertainty,cheng2009uncertainty}). This is pertinent in operational context when model predictions guide decision-making processes and uncertainty propagation and data assimilation (\citet{TerejanuP_IF_2007, TerejanuP_CBD_2008}) need to be performed in real-time. StackedGP is especially applicable in the context of GP emulators driven by forcing variables predicted by other GP or StackedGP models. A simple 2D puff advection example is provided to showcase StackedGP's applicability in uncertainty propagation using emulated dynamical systems.

The paper is organized as follows. Section~\ref{sec:background} provides a brief introduction to GP. Section~\ref{sec:uncertainGP} re-derives the expectations of a GP with uncertain inputs for squared exponential kernel, and provides a novel derivation for the polynomial kernel. Section~\ref{sec:StackedGP} generalizes the StackedGP to an arbitrary number of layers and nodes, and discusses the advantages and limitations of the proposed model. Four numerical results are presented in Section~\ref{sec:results} and conclusions are given in Section~\ref{sec:conclusions}.

\section{Gaussian Process Background}
\label{sec:background}

Unlike parametric models, non-parametric models provide infinite dimensional parameters for modeling the distribution of the data. Gaussian processes are popular non-parametric models that have many uses in  machine learning (\citet{rasmussen2006gaussian,williams1996gaussian,gpwilliams1998prediction,reggente2014prediction}) and environmental modeling as previously described. 

Given $D=\{\bX,\bz\}$, a set of $n$ data points, each consisting of $d$ inputs ($\bX \in \Re^{n \times d}$) and one output ($\bz \in \Re^{n}$), the output of the $i$th data point, $z_i$, is modeled as follows:
\begin{align}
z_i&=g(\bx_i)+\epsilon_i^z \\
\epsilon_i^z &\sim N(0,\sigma^2_{\epsilon_z}) \\
g &\sim \textrm{GP}(0, k_z(\cdot,\cdot))
\end{align}

Here, $g$ represents a latent function with zero mean Gaussian process prior and kernel or covariance function $k_z(\cdot,\cdot)$. The kernel measures the similarity between two inputs, $\bx_i$ and $\bx_j$. For example, the squared exponential or radial basis function (RBF) kernel is defined as follows.
\begin{equation}
k_z(\bx_i,\bx_i) = \phi \exp\left\{-\theta \|\bx_i-\bx_j\|_2 \right\}
\end{equation}

The hyperparemeters, $\sigma^2_{\epsilon_z}$, and e.g. $\phi$ and $\theta$ corresponding to the RBF kernel, are estimated using the maximum likelihood approach, where the log-likelihood is given by,
\begin{align}
	\textrm{ln} p(\bz|\bX,\phi,\theta,\sigma^2_{\epsilon_z}) = -\frac{1}{2}\bz^T (\bK_z+\sigma^2_{\epsilon_z}I)\bz - \frac{1}{2}\textrm{ln}(\bK_z+\sigma^2_{\epsilon_z}I) - \frac{n}{2}\textrm{ln}2\pi~,
\end{align}
and the covariance matrix $\bK_z$ is an $n \times n$ Gram matrix with elements $K_{ij}=k_z(\bx_i,\bx_i)$.

Once the hyperparameters are estimated, the predictive distribution of $z^*$ at a new testing input $\bx^*$, is given by the following normal distribution.
\begin{align}
z^* &\sim N\left(\mu_{z^*},\sigma_{z^*}^2\right) \\
\mu_{z^*} &= \bk_z^T\bC_z^{-1}\bz \\
\sigma^2_{z^*} &= k_z\left(\bx^*,\bx^*\right)+\sigma^2_{z}-\bk_z^T\bC_z^{-1}\bk_z \\
\bC_z &= \bK_z+\sigma^2_{\epsilon_z}I
\end{align}

In the following section we provide the background for a simple StackedGP as an extension to GP with uncertain inputs as initially developed by \citet{girard2002gaussian}.

\section{Simple StackedGP - Two Chained Gaussian Processes}
\label{sec:uncertainGP}

\begin{figure}[H]
	\centering
	\input{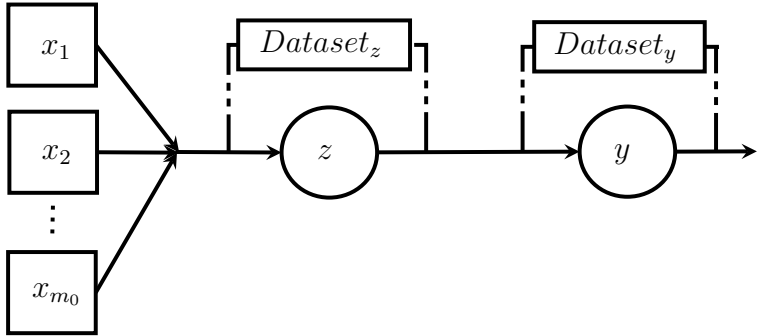}
	\caption{Simple StackedGP - two chained Gaussian processes. Circles represent a GP node and squares represent the observable inputs. $Dataset_z$ and $Dataset_y$ are used to train the first and second GP, respectively.}
	\label{fig:stackedGPsimplecase}
\end{figure}

Consider the following simple StackedGP in Fig.~\ref{fig:stackedGPsimplecase} given by two chained GPs with their own training dataset. The input to the first GP is given by the vector $\bx$. The output of the first GP, $z$ governs the input to the second GP, and $y$ is the final output of the StackedGP in Fig.~\ref{fig:stackedGPsimplecase}.

The goal of this section is to introduce the mechanism of obtaining analytical expectations of two-layer StackedGPs for both RBF and polynomial kernels. Note that the predictive distribution of even a simple StackedGP as the one in Fig.~\ref{fig:stackedGPsimplecase} is non-Gaussian, however its mean and variance can be obtained analytically. In the next section we will generalize the approach to obtain the approximate expectations of StackedGPs with arbitrary number of layers and nodes per layer.

We start with providing analytical expressions for mean and variance for a general kernel, and follow with specific expressions for RBF kernel as initially derived by \citet{girard2002gaussian}, and then with a novel derivation for polynomial kernel.

The predicted mean of the StackedGP with input $\bx^*$ is obtained using the law of total expectation by integrating out the intermediate variable $z^*$:
\begin{equation}
E[y^*|\by, \bx^*] = \E_{z^*}\left\{\E[y^*|\by, \bx^*, z^*]\right\}
\end{equation}

Here, $\E[y^*|\by, \bx^*, z^*]=\bk_{y}^T\bC_{y}^{-1}\by$ is the expectation of a standard GP with input $z$ and output $y$, and it can be expanded as follows:
\begin{align}
\E[y^*|\by, \bx^*, z^*]=\by^T \sum_{i=1}^n \bC_{y}^{-1}(,i) k_{y}\left(z^*,z_i\right)\label{eq:kca}~,
\end{align}
where $\bC_{y}$ is the covariance matrix of the second GP and $k_{y}(z^{*},z_{i})$ is the kernel between the predicted variable $z^*$ and the $i^{th}$ training data point $z_{i}$, and $n$ is the number of training points for the target node. The final predicted analytical mean of $y^*$ can be written as
\begin{equation}
\E[y^*|\by,\bx^*]=\by^T \sum_{i=1}^n \bC_{y}^{-1}(,i) \underbrace{\E_{z^*}\left\{k_{y}(z^{*},z_{i})\right\}}_{\Delta_1}~.
\label{equ:simplepredictedmean}
\end{equation}
$\E_{z^*}[k_{y}(z^{*},z_i)]$ is the key integration to obtain the analytical predicted mean. The expectation in Eq.~\ref{equ:simplepredictedmean} is with respect to a normal distribution with mean $\mu_{z^*}$ and variance $\sigma^2_{z^*}$ as obtained from the prediction of the first GP. The expectation can be obtained analytically for RBF and polynomial kernels as shown in the following two subsections. 

The variance of the StackedGP can be obtained similarly using the law of total variance.

\small
\begin{align}
	\begin{split}
		\Var\left(y^*|\by, \bx^*\right)
		=&\E_{z^*}\left[\Var\left(y^*|\by, \bx^*, z^*\right)\right]+\Var_{z^*}\left(\E\left[y^*|\by, \bx^*, z^*\right] \right)\\
		=&\E_{z^*}\left[k_{y}(z^{*},z^{*})+\sigma_{y}^2-\bk_{y}^T\bC_{y}^{-1}\bk_{y}\right]
		+\Var_{z^*}\left( \bk_{y}^T\bC_{y}^{-1}\by\right)\\
		=& \sigma_{\varepsilon_y}^2 + \underbrace{\E_{z^*}\left[k_{y}(z^{*},z^{*})\right]}_{\Delta_2} -\E_{z^*}\left[\bk_{y}^T\bC_{y}^{-1}\bk_{y}\right]
		+\Var_{z^*}\left( \bk_{y}^T\bC_{y}^{-1}\by\right)\\
	\end{split}
	\label{equ:simplepredictedvar}
\end{align}
\normalsize 

Here, $\sigma_{\varepsilon_y}^2$ is the noise variance of the target GP and $\E_{z^*}\left[\bk_{y}^T\bC_{y}^{-1}\bk_{y}\right]$ can be obtained using the following expansion.
%
\begin{align}
	\E_{z^*}\left[\bk_{y}^T\bC_{y}^{-1}\bk_{y}\right] =\sum_{i=1}^n \sum_{j=1}^n \bC_{y}^{-1}(i,j) \underbrace{E_{z^*}\left[k_{y}\left(z^*,z_{i})\right)
	k_{y}\left(z^*,z_{j}) \right)\right]}_{\Delta_3}
	\label{equ:generalDelta2}
\end{align}

The last term in Eq.~\ref{equ:simplepredictedvar} is given by,
\begin{align}
\Var_{z^*}\left( \bk_{y}^T\bC_{y}^{-1}\by\right) &= \bY^T \bC_{y}^{-1}\bSigma_k\bC_{y}^{-1}\bY \label{equ:variance_term}
\end{align}
where, $\bSigma_k = \Var_{z^*}\left(\bk_{y}\right) \in \Re^{n \times n}$ can be expressed as 
\begin{align}
	\bSigma_k = \E_{z*}\left[\bk_{y} \bk_{y}^T\right]-\E_{z^*}\left[\bk_{y} \right]\E_{z^*}\left[ \bk_{y}'\right].
	\label{equ:Sigmak}
\end{align}

Note that $\bSigma_k$ is computed using the two integrations of $\Delta_1$ and $\Delta_3$. 

In the following two subsections, we will provide the analytical first and second moments of StackedGP for RBF and polynomial kernels.



\subsection{RBF Kernel - Simple Case}

Using the RBF kernel $k_{y}(z^*,z_i)=\phi\exp \left\{-\theta(z^*-z_i)^2\right\}$ to evaluate $\Delta_1$ in Eq.~\ref{equ:simplepredictedmean} we obtain:

\small
\begin{align*}
&\Delta_1=\sqrt{\frac{(1/(2\theta))}{\sigma_{z^*}^2+(1/(2\theta))}}
\exp \left\{-\frac{(z_{i}-\mu_{z^*})^2}{2(\sigma^2_{z^*}+1/(2\theta))}\right\}
\end{align*}
\begin{align}
		&\E[y^*|\by,\bx^*]=\phi\by^T \sqrt{\frac{(1/(2\theta))}{\sigma_{z^*}^2+(1/(2\theta))}}
		\times\sum_{i=1}^n\bC_{y}^{-1}(,i)\exp \left\{-\frac{(z_{i}-\mu_{z^*})^2}{2(\sigma^2_{z^*}+1/(2\theta))}\right\}
	\label{equ:RBFsimplepredictedmean}
\end{align}
\normalsize

Here, $\theta$ is the corresponding length scale in the target node, $\phi$ is the kernel's variance, and $\by^T$ is the output training points that have been used during training of the target GP node. 

For RBF kernel, $\Delta_2 = \phi$ and $\Delta_3$ in Eq.~\ref{equ:simplepredictedvar} can be calculated using the following expression. 
\begin{align*}
\Delta_3
=\phi^2 \sqrt{\frac{1/(4\theta)}{1/(4\theta)+\sigma_{z^*}^2}}
&\times\exp\left\{-\frac{\theta(z_{i}-z_{j})^2}{2}-\frac{\left[(z_{i}+z_{j})/2-\mu_{z^*}\right]^2}{2\left(1/(4\theta)+\sigma_{z^*}^2\right)}\right\}
\end{align*}
Here, $z_{i}$ is the $i^{th}$ input training data point for the target node. Finally, the predicted variance is given by:

\small
\begin{align}
	\begin{split}
		\Var\left(y^*|\by,\bx^*\right)
		=&\sigma_{\varepsilon_y}^2+\phi+\by^T \bC_{y}^{-1}\bSigma_k\bC_{y}^{-1}\by
		-\phi^2\sum_{i=1}^n \sum_{j=1}^n \bC_{y}^{-1}(i,j) \sqrt{\frac{1/(4\theta)}{1/(4\theta)+\sigma_{z^*}^2}}\\
		&\times\exp\left\{-\frac{\theta(z_{i}-z_{j})^2}{2}-\frac{\left[(z_{i}+z_{j})/2-\mu_{z^*}\right]^2}{2\left(1/(4\theta)+\sigma_{z^*}^2\right)}\right\}\label{equ:RBFsimplePredictedVar}
	\end{split}
\end{align}
\normalsize

These analytical expressions corresponding to the RBF kernel coincide with those derived by \citet{girard2002gaussian} and \citet{candela2003propagation}. We have provided them here for completeness and to emphasize the role of uncertainty in the network as described in the following sections. In the next subsection we provide novel analytical expressions for the predicted mean and variance of StackedGP when using polynomial kernels.

%

\subsection{Polynomial Kernel - Simple Case}

Following the same simple StackedGP configuration and a $d$-order polynomial kernel at the target node $k_{y}\left(z^*,z_i\right)=(z^**z_i)^d$, the predicted mean of Eq.~\ref{equ:simplepredictedmean} can be calculated as 
\begin{align*}
\E[y^*|\by,\bx^*]=\by^T \sum_{i=1}^n \bC_{y}^{-1}(,i)(a_d z_i^d)
\end{align*}
where $\Delta_1 = (a_d z_i^d)$ and $a_d$ follows the non-central moments of the normal distribution, namely
\begin{align}
a_d = \displaystyle \sum_{u=0}^{\floor{\frac{d}{2}}}{\dbinom{d}{2u}(2u-1)!!\sigma^{2u}_{z^*} \mu^{d - 2u}_{z^*}}.
\label{equ:noncentralmoments}
\end{align}

The expression for the predicted variance in Eq.~\ref{equ:simplepredictedvar}  is obtained by substituting $\Delta_2=a_{2d}$ and $\Delta_3=a_{2d} z_i^dz_j^d$ where $a_{2d}$ is calculated using Eq.~\ref{equ:noncentralmoments}. Finally, the predicted variance in the case of polynomial kernel is given by,
\begin{align*}
\Var_{poly}\left(y^*|\by,\bx^*\right)
= \sigma_{\varepsilon_y}^2 +a_{2d}+ \by^T \bC_{y}^{-1}\bSigma_{k}\bC_{y}^{-1}\by - \sum_{i=1}^n \sum_{j=1}^n a_{2d} z_i^dz_j^d\bC_{y}^{-1}(i,j).
\end{align*}


\section{Stacked Gaussian Process - Generalization}
\label{sec:StackedGP}

The goal of this section is to extend the previous StackedGP to an arbitrary number of layers and nodes per layer. First, we start by presenting the analytical mean and variance of a two-layer StackedGP with arbitrary number of nodes in the first layer. Second, we provide a discussion on accommodating an arbitrary number of output nodes in the second layer. Finally, we present an algorithm to compute the approximate mean and variance of a generalized StackedGP, and discuss the advantages and limitations of the model.

\subsection{Generalized Number of Nodes in the First Layer of a Two Layer StackedGP}

\begin{figure}[H]
	\centering
	\input{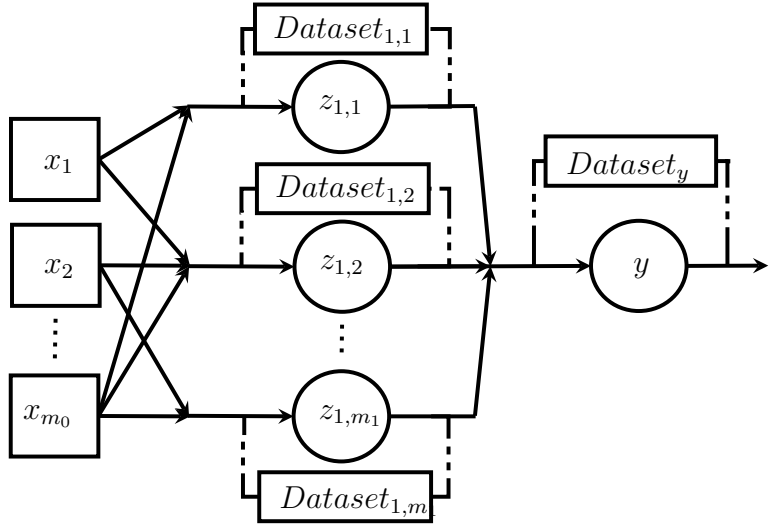}
	\caption{StackedGP with multiple nodes in the first layer. 
	Circles represent GP nodes and squares represent the observable inputs.}
	\label{fig:stackedGPGenerInputs}
\end{figure}

Consider an arbitrary number of nodes in the first layer as an extension of the simple two layer StackedGP in the previous section while keeping the single output, see Fig.~\ref{fig:stackedGPGenerInputs}. The analytical expectations presented here will require the independence assumption for the input uncertainties in the target node. Namely, the outputs of the first layer, $z_{1,1}, z_{1,2} \ldots z_{1,m_1}$ are considered independent. In this context, the predicted mean and variance of the target node can be generalized as follows:
 \begin{align}
 	\E[y^*|\by, \bx^*] = \bv^T\bC_{y}^{-1}\by
 	 \label{equ:generalPredictedmean}
 \end{align}
 \begin{equation}
 var[y^*] = \sigma_{\varepsilon_y}^2 + \Delta_{2_g} + \underbrace{\by^T \bC^{-1} \bSigma_k \bC^{-1} \by}_{\zeta}  - \sum_{n,n} (\bC^{-1}\odot \bH)
 \label{equ:generalPredictedVar}
 \end{equation}

Here, the symbol "$\odot$" is used for element-wise product or Hadamard product. The elements of the vector $\bv \in \Re^{n \times 1}$ act as kernels under the uncertain inputs. Scalers $\Delta_{2_g}$ and $\zeta$ (third term in Eq.~\ref{equ:generalPredictedVar}), as well as $\bH \in \Re^{n \times n}$ reflect integrations of kernel functions under the uncertain inputs as shown in the following two subsections for the RBF and the polynomial kernel. 


\subsubsection{RBF kernel - Generalized Number of Nodes in the First Layer}
\label{sec:RBFkernel}

The analytical mean in the case of the RBF kernel for the output node is obtained using the following elements of the $\bv$ vector in Eq.~\ref{equ:generalPredictedmean}.
%
%
\begin{align}
	v_i&=w q_i \\
	w &= \prod_{j=1}^{m_1} \sqrt{\frac{1/(2\theta_j)}{((1/(2\theta_j)+\sigma_{z_{j}^*}^2)}}\\
	q_{i} &= \phi \exp \left\{\sum_{j=1}^{m_1} -\frac{(z_{j_i}-\mu_{z_{j}^*})^2}{2((1/(2\theta_j)+\sigma_{z_{j}^*}^2)}\right\} 
	\label{equ:generalizedRBFmean}
\end{align}
 
Here, $i = 1..n$, where $n$ is the number of training data points for the output node, $m_1$ is the number of inputs to the output GP node, and $z_{j_i}$ is the $j^{th}$ element of the $i^{th}$ training data point. Note that the predicted mean of the StackedGP has the same form as the standard GP but with two main differences. First, the kernel evaluations $v_i$ measure the similarity between the $i^{th}$ training data and the predicted mean $\bmu_{z^*}$ from the previous layer instead of the direct input. Second, the similarity is discounted based on the input uncertainty $\sigma_{z_{j}^*}^2$. Note, that if we set $\sigma_{z_{j}^*}^2$ to zero, we obtain a common product of RBF kernels corresponding to each node in the first layer. However, the larger the input uncertainty for a particular node the lower the similarity on that particular dimension.


To obtain the analytical variance for the RBF kernel in Eq.~\ref{equ:generalPredictedVar}, we use the following relations:  $\Delta_{2_g}=\phi$ and $\bH = u \bP$ where the scalar $u$ and the elements of $\bP \in \Re^{n \times n}$ are given by
\begin{align}
	u = \prod_{j=1}^{m_1} \sqrt{\frac{1/(4\theta_j)}{((1/(4\theta_j)+\sigma_{z_{j}^*}^2)}}
	\label{equ:generalizedRBvarU}
\end{align}
\begin{equation}
	\bP_{a,b}=\phi^2 \exp\left\{- \sum\limits_{j=1}^{m_1}\left\{ \frac{\theta_j(z_{j_a}-z_{j_{b}})^2}{2} \right.\right. 
	\left.\left.+\frac{\left[(z_{j_a}+z_{j_{b}})/2-\mu_{z_{j}^*}\right]^2}{2\left(1/(4\theta_j)+\sigma_{z_{j}^*}^2\right)}\right\}\right\}~.
	\label{equ:generalizedRBvarP}
\end{equation}

Using Eq.~\ref{equ:Sigmak}, we can get the following expression for $\bSigma_k$: 
\begin{equation}
\bSigma_k = u \bP - w^2 \bT
\end{equation}
where the elements of the matrix $\bT\in \Re^{n \times n}$ are defined as
\begin{equation}
	\bT_{a,b} = \phi^2 \exp \left\{-\sum\limits_{j=1}^{m_1} \frac{(z_{j_a}-\mu_{z_{j}^*})^2 + (z_{j_{b}}-\mu_{z_{j}^*})^2}{2(1/(2\theta_j)+\sigma_{z_{j}^*}^2)}\right\}~.
	\label{equ:generalizedRBvarT}
\end{equation}

Note that if the uncertainty from the first layer $\sigma_{z_{j}^*}^2=0$ then we obtain the same standard variance of a Gaussian process. Namely, the scalers $u$ and $w$ become one and $\bP_{a,b}=\bT_{a,b}=\bk_y(z_{a},\mu_{z_{j}^*})\bk_y(z_{{b}},\mu_{z_{j}^*})^T$, which yields $\bSigma_k = \bm{0}$ and thus $\zeta=0$ in Eq.~\ref{equ:generalPredictedVar}. As a result, in the case of certain inputs, the predicted variance of the StackedGP is similar to the standard GP, namely $\sigma_{\varepsilon_y}^2 + \phi - \bk_{y}^T \bC^{-1}\bk_y$. Here, $\bk_y$ is the kernel evaluated at the training point and the predicted mean of the first layer. In other words, if we have  certain inputs, we get standard GP prediction. Otherwise, the uncertainty in the first layer is propagated to the second layer, increasing the predictive uncertainty of the StackedGP output. 

In the next section we expand these derivations to polynomial kernels.

\subsubsection{Polynomial Kernel - Generalized Number of Nodes in the First Layer}

The analytical mean in the case of polynomial kernel of order $d$ for the output node is obtained using the following multinomial expansion for the $i^{th}$ element of the $\bv$ vector in Eq.~\ref{equ:generalPredictedmean}.
\begin{align}
	\bv_i = \displaystyle \sum_{p_1+p_2+...p_{m_1} = d}{\dbinom{d}{p_1,p_2,...p_{m_1}} \displaystyle \prod_{1\leqslant t \leqslant {m_1}}[a_{p_{t}}z_{t_i}^{p_t}]}~.
	\label{equ:generalizedPolymean}
\end{align}

Here, $p_i$ indicates the power of the $t^{th}$ input with $1\leqslant t \leqslant {m_1}$. In additions, $\dbinom{d}{p_1,p_2,...p_{m_1}} = \frac{d!}{p_1!p_2!...p_{m_1}!}$, and the coefficient $a_{p_{t}}$ follows the non-central moment of the normal distribution shown in Eq.~\ref{equ:noncentralmoments}. Note, that in the absence of input uncertainty, namely setting $\sigma_{z_{j}^*}^2=0$, we actually set all but the first term in Eq.~\ref{equ:noncentralmoments} to zero, which results in the same formula for the mean of a standard GP with a polynomial kernel of order $d$.

To obtain the analytical variance for the polynomial kernel in Eq.~\ref{equ:generalPredictedVar}, we use the following relations:
\begin{align}
\Delta_{2_g} &= \displaystyle \sum_{p_1+p_2+...p_{m_1} = d}{\dbinom{d}{p_1,p_2,...p_{m_1}} \displaystyle \prod_{1\leqslant t \leqslant {m_1}}[a_{2p_{t}}]}
\label{equ:generalizedPolyDelta} \\
a_{2p_{t}} &= \sum_{u=0}^{\floor{\frac{2*p_t}{2}}}{\dbinom{2*p_{t}}{2u}(2u-1)!!\sigma^{2u}_{z_t^*} \mu^{2*p_t - 2u}_{z_t^*}}~.
\end{align}

Using Eq.~\ref{equ:Sigmak}, we can get the expression for $\bSigma_k$: 
\begin{equation}
\bSigma_k = \bH - \bv\bv^T
\end{equation}
where the elements of the matrix $\bH\in \Re^{n \times n}$ are obtained using the following multinomial expansion,
\begin{align}
	\bH_{i,j} &= \displaystyle \sum_{p_1+...p_{m_1} = d}\sum_{q_1+...q_{m_1} = d}{\dbinom{d}{p_1,...p_{m_1}} \dbinom{d}{q_1,...q_{m_1}} \displaystyle \prod_{1\leqslant t \leqslant {m_1}}[a_{p_{t},q_{t}}z_{t_i}^{p_t}z_{t_j}^{q_t}]}
	\label{equ:generalizedPolyvarH} \\
	a_{p_{t},q_{t}} &= \sum_{u=0}^{\floor{\frac{p_t+q_t}{2}}}{\dbinom{p_{t}+q_t}{2u}(2u-1)!!\sigma^{2u}_{z_t^*} \mu^{p_t+q_t - 2u}_{z_t^*}}
\end{align}

Similarly as in the RBF case, if there is no uncertainty coming from the first layer, namely $\sigma_{z_{j}^*}^2=0$, then $\bH = \bv\bv^T$, which yields $\bSigma_k = \bm{0}$ and thus $\zeta=0$ in Eq.~\ref{equ:generalPredictedVar}. Since $\bH_{a,b}=\bk_y(z_{a},\mu_{z_{j}^*})\bk_y(z_{{b}},\mu_{z_{j}^*})^T$, this leads to a predicted variance of the StackedGP similar to the standard GP with polynomial kernel, $\sigma_{\varepsilon_y}^2 + \Delta_{2_g} - \bk_{y}^T \bC^{-1}\bk_y$. Here, $\bk_y$ is the polynomial kernel evaluated at the predicted mean of the first layer and the training points. 


Note that the first two moments can be easily obtained also for kernels that involve sums of RBF and polynomial kernels. In the following section we discuss how we can expand the two-layer network to arbitrary number of outputs, and finally the assumptions needed to obtain approximate expectations in a StackedGP with arbitrary number of layers and nodes per layer.

\subsection{StackedGP with Arbitrary Number of Layers and Nodes per Layer}

The only assumption in the previous sections is that the outputs of layers that propagate as inputs to the next layer are independent. This applies also to the extension of the previous StackedGP to an arbitrary number of outputs in the last layer. This assumption is for convenience as the derivations are significantly more involving, however the methodology can accommodate correlated inputs. For example, co-kriging methods (\citet{cressie1992statistics}) and dependent GPs (\citet{boyle2005dependent}) provide an alternative formulation for obtaining coupled outputs. Any of these models might be used to generate correlated outputs for any layer, however these correlations need to be incorporated into the StackedGP expectations. In our numerical results, we have opted to pre-process the training data using independent component analysis (ICA) to obtain independent projections that are finally used to train the GPs. Note that this procedure does not include the deterministic input observations. We plan to extend the derivations to account for correlations in our next study.

\begin{figure}[H]
	\centering
	\input{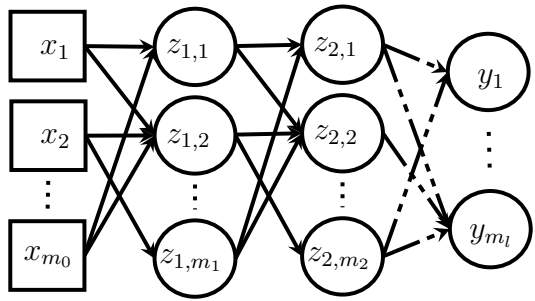}
	\caption{Stacked Gaussian Process model. The output dimension of $\by(x)$ is $m_l$ where the model has $l$ stacked layers and each layer has $m_i$  GP nodes ($i$ refers to the index of the layer). Circles represent a GP node and squares represent the observable inputs.}
	\label{fig:StackedGP}
\end{figure}

The objective of this section is to build a StackedGP to model an $m_l$ dimensional function $\by(x)$ as shown in Fig.~\ref{fig:StackedGP}. The model has $l$ stacked layers with each layer having $m_i$ GP nodes ($l$ refers to the index of the layer and the value of $m_l$ can be different from layer to layer). We assume that we are given the following set of training datasets $D_{train} = \{D_1,D_2,...D_{Q}\}$, where $Q=\sum_{i=1}^{l}m_i$ represents the total number of nodes in the model. In this stacked model each node is independently trained using its own available dataset $D_q$, where $q=1..Q$. Thus, each node acts as a standalone standard GP, where the hyper-parameter optimization/inference is conducted using node specific datasets. 

While for two-layer StackedGP the mean and the variance can be obtained analytical for both RBF and polynomial kernel, in the case of three or more layers the expectations are intractable for the RBF kernel, and in the case of polynomial kernels, they involve keeping track of large number of terms. We have opted to approximately propagate the uncertainty from layer to layer and approximate the expectations of the StackedGP. Note that even if we are able to obtain analytical expectations for a chain of two GPs, the underlying distribution is still non-Gaussian. As a result, in addition to the independence assumption for the outputs of each layer, we add another assumption which involves approximating the distribution of the output of each layer with a Gaussian distribution. Given the analytical mean and variance, we use the maximum entropy principle to obtain the Gaussian approximation. The effect of this approximation is an increase in the uncertainty that is propagated through the network, resulting in conservative predictions. 

In large networks or multi-step predictions this uncertainty inflation due to maximum entropy approach might have a significant impact. However, this impact is minimized in applications such as data assimilation, where frequent measurements can reduce the predicted uncertainty. Furthermore, a sensitivity analysis can be used to determine the nodes and the inputs that contribute the most to the final uncertainty of the quantity of interest. This way, one can allocate resources such as targeted data collection or kernel tunning to improve the GP model of the node with the highest uncertainty contribution.

Finally, Eqs.~\ref{equ:generalPredictedmean} and \ref{equ:generalPredictedVar} provide the main mechanism to obtain the approximate mean and variance of a layer given the predictions of the previous layer. This process is applied sequentially until the mean and variance of the final quantities of interest are obtained. Algorithm~\ref{algo:stackedGP} demonstrates how a general StackedGP is built and the steps required to obtain the desired expectations. We have built a Python package based on \citet{gpy2014} to create general StackedGP models, perform optimizations and calculate predictions. The software package will be available at the time of publication under an open source license.

 \begin{algorithm}[H]
 	\caption{StackedGP - model building and uncertainty propagation}
 	\label{algo:stackedGP}
 	\begin{algorithmic}[1]
 		\REQUIRE $D_{train}=[D_1,D_2,...D_Q]$. $Q$ number of nodes in the StackedGP.
 		\REQUIRE $nodeLayerIdx=\{(l,n)_j\}_{j=1\ldots Q}$. $Q$ tuples of layer and node index for each node. 
 		\REQUIRE $stackedStructure$: an array of $Q$ lists, where each list $stackedStructure[node]$ has an arbitrary number of tuples to specify the inputs nodes to the current $node$. 
 		\REQUIRE New observation $\bx^*$\\
 		\COMMENT{\# Create StackedGP}
 		\FOR{$i~in~range(1,Q):$} 
 			\STATE kernel initialization (RBF, Polynomial, or RBF + Polynomial).
 			\IF{$nodeLayerIdx[i][1] !=0$} 
 				\STATE apply ICA on $D_{train}[i].X$.
 			\ENDIF
 			\STATE init $node$ with inputs $D_{train}[i].X$ and outputs $D_{train}[i].Y$
 			\STATE estimate hyperparameters for $node$.
 			\STATE add $node$ to StackedGP at location $nodeLayerIdx[i]$
 		\ENDFOR \\
 		\COMMENT{\# Uncertainty propagation}
 		\FOR{$i~in~range(number~of~layers)$}
 			\FOR{$node~in~layer[i].nodes$}
 				\STATE extract mean and variance of all inputs from $stackedStructure[node]$
 				\COMMENT{\# Calculate the mean and variance for the current node}
 				\STATE RBF kernel: mean (Eqs.~\ref{equ:generalPredictedmean}, \ref{equ:generalizedRBFmean}), and variance (Eqs.~\ref{equ:generalPredictedVar}, \ref{equ:generalizedRBvarU}, \ref{equ:generalizedRBvarP}, \ref{equ:generalizedRBvarT}).
 				\STATE Polynomial kernel: mean (Eqs.~\ref{equ:generalPredictedmean}, \ref{equ:generalizedPolymean}), variance (Eqs.~\ref{equ:generalPredictedVar}, \ref{equ:generalizedPolyDelta}, \ref{equ:generalizedPolyvarH}).
 			\ENDFOR
 		\ENDFOR
 	\end{algorithmic}
 \end{algorithm} 

One limitation of the model is related to the matrix inversion required by the standard GP model, which takes $\mathcal{O}(n^3)$ operations, where $n$ is the number of training data points for a particular node. Several approaches have been proposed to deal with the curse of dimensionality: kernel mixing (\citet{Higdon1998}), sparse GP with pseudo-inputs (\citet{snelson2006sparse}), incremental local Gaussian regression (\citet{meier2014incremental}), and inversion free approaches (\citet{doi:10.1080/10618600.2016.1164056}).

When the output of various layers is high-dimensional, then dimensionality reduction techniques can be added to pre-process the training data (\citet{10.2307/27640080}). Also, various operations in Algorithm~\ref{algo:stackedGP} are easily parallelizable. Namely, the optimization for hyperparameter estimation of each node can be carried out in parallel as well as within layer propagation of information from the previous layer. Obviously, this  computational efficiency over multi-output methods comes at a cost of properly accommodating for the correlation of the outputs. 

\section{Numerical Results}
\label{sec:results}
In this section we provide four different examples to demonstrate the applicability of StackedGP. The first example involves the use of a set of synthetic datasets to demonstrate that StackedGP is able to capture the outputs of simple composite functions. In the second example, we use StackedGP to combine two real datasets to predict the burned area as part of a forest fire application. The third application corresponds to the Jura geological dataset, where the StackedGP is used to enhance the prediction of a primary response using intermediate predictions of secondary responses. Finally, we demonstrate the use of StackedGP in the context of emulated dynamical systems for 2D puff advection driven by uncertain inputs for multi-step predictions.

\subsection{Model Composition - Synthetic Datasets}
\begin{figure}[htp!]
	\centering
	\input{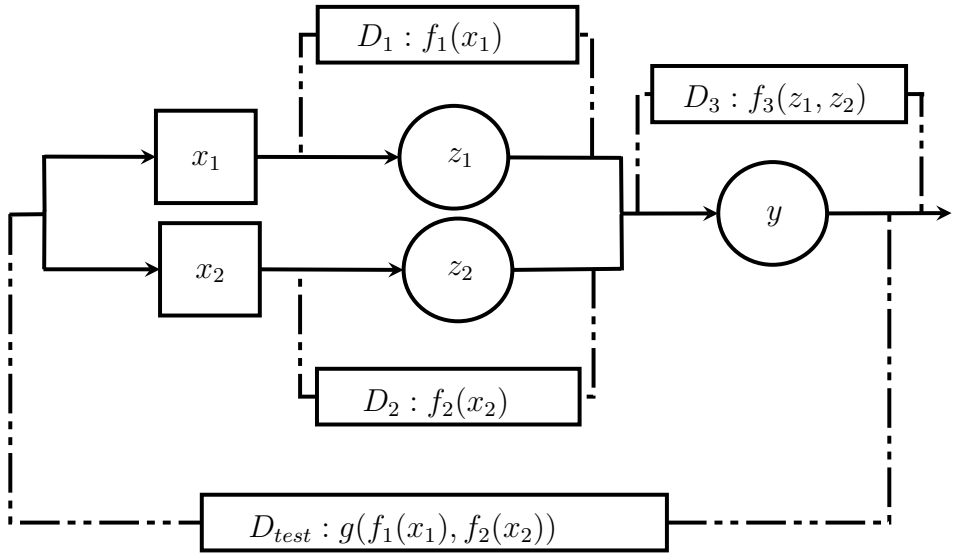}
	\caption{Example 1 (synthetic datasets) - StackedGP for predicting the output of a composite function. The input to this model are $x_1$ and $x_2$, and the final output is $y$ with $z_1$ and $z_2$ as outputs from the middle layer. We use three datasets [$D_1$, $D_2$, and $D_3$] for training and $D_{test}$ for testing. These datasets are shown in Table~\ref{tab:toyexamples}.}
	\label{fig:toy_graph} 
\end{figure}

\begin{figure*}[h]
	\begin{minipage}{0.50\linewidth}
		\centering
		
		\begin{tabular}{c c c c c c }
			\begin{minipage}{.2\textwidth}
				\small
				\textbf{Model}
			\end{minipage} 
			&\begin{minipage}{.35\textwidth}
				\tiny
				$z_i = x_i^2$\\
				$y = {z_1 + 2*z_2}$		
			\end{minipage} 
			
			&\begin{minipage}{.4\textwidth}
				\tiny
				$z_1 = \ln(x_1)$\\
				$z_2 = \ln(x_2^3)$\\
				$y = sin(\sqrt{z_1 + z_2})$
			\end{minipage} 
			&\begin{minipage}{.35\textwidth}
				\tiny
				$z_1 = sin(x_1)$\\
				$z_2 = sin(x_2)$\\
				$y = z_1*z_2$
			\end{minipage}
			&\begin{minipage}{.45\textwidth}
				\tiny
				$z_i = x_i^2$\\
				$y = \sqrt{(z_1+z_2)}+3*cos(\sqrt{z_1+z_2})+5$
			\end{minipage}  \\
			\begin{minipage}{.2\textwidth}
				\small
				\textbf{Training}
			\end{minipage} 
			&\begin{minipage}{.35\textwidth}
				\includegraphics[scale=0.25]{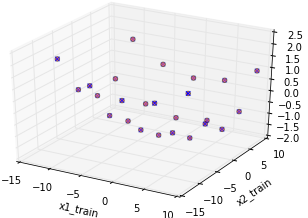}
			\end{minipage} 
			&\begin{minipage}{.35\textwidth}
				\includegraphics[scale=0.25]{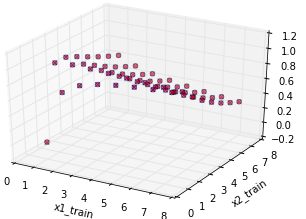}
			\end{minipage} 
			&\begin{minipage}{.35\textwidth}
				\includegraphics[scale=0.25]{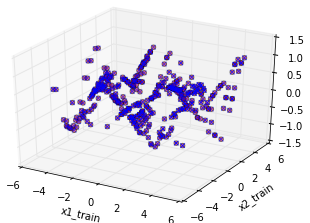}
			\end{minipage}
			&\begin{minipage}{.35\textwidth}
				\includegraphics[scale=0.25]{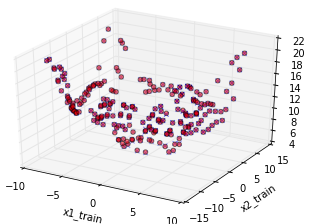}
			\end{minipage}  \\
			\begin{minipage}{.2\textwidth}
				\small
				\textbf{Testing}
			\end{minipage} 
			&\begin{minipage}{.35\textwidth}
				\includegraphics[scale=0.25]{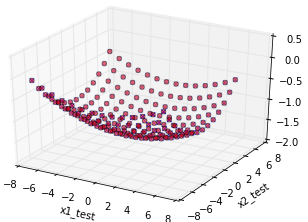}
			\end{minipage} 
			
			&\begin{minipage}{.35\textwidth}
				\includegraphics[scale=0.25]{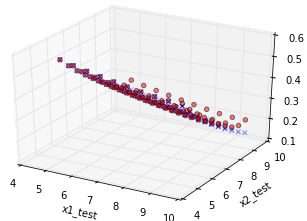}
			\end{minipage} 
			&\begin{minipage}{.35\textwidth}
				\includegraphics[scale=0.25]{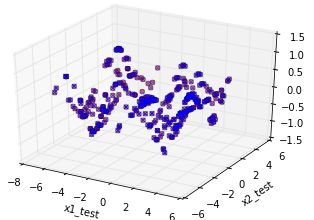}
			\end{minipage}
			&\begin{minipage}{.35\textwidth}
				\includegraphics[scale=0.25]{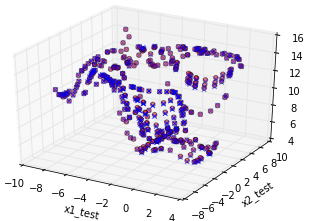}
			\end{minipage} \\ \\
			
			\begin{minipage}{.2\textwidth}
				\small
				\vspace{0.2in} \textbf{RMSE \\ AvgRatio }
			\end{minipage} 
			&\begin{minipage}{.2\textwidth}
				\centering
				$0.0007$ \\$0.15$
			\end{minipage} 
			&\begin{minipage}{.2\textwidth}
				\centering
				$0.040$  \\$0.17$
			\end{minipage} 
			&\begin{minipage}{.2\textwidth}
				\centering
				$0.0067$  \\$0.10$
			\end{minipage}
			&\begin{minipage}{.2\textwidth}
				\centering
				$0.33$\\  $0.41$
			\end{minipage} \\
		\end{tabular}
	\end{minipage}
	\captionof{table}{Example 1 (synthetic datasets) - Applying the stacked model shown in Figure~\ref{fig:toy_graph} on different synthetic  scenarios. These figures show the training set and the final predictions for the input $x_1$ and $x_2$. In all figures actual data is represented with a blue 'x', and the predicted mean with a red dot.}
	\label{tab:toyexamples} 
\end{figure*}

StackedGPs are build using synthetically generated data from four composite functions as shown in Table~\ref{tab:toyexamples}. One set of data is generated for the mappings between $(x_1,x_2)$ and $(z_1,z_2)$, and another data set is generated for the mappings between $(z_1,z_2)$ and $y$. The two datasets are used to build three independent GPs,  which are then stacked to obtain the StackedGP shown in Figure~\ref{fig:toy_graph}. Table~\ref{tab:toyexamples} shows the training set and the testing set for each scenario, as well as the ability of the StackedGP to capture different non-linear hierarchical functions. 

The root mean square error (RMSE) is used to measure the performance of the stacked model by comparing the prediction of the StackedGP at various inputs $(x_1,x_2)$ and the true value $y$ of the composite function at those inputs. In addition, the average ratio $\frac{|\hat{y}-y^*|}{\sigma}$ is reported to verify that the true value falls within the $95\%$ credible interval as predicted by the model. This corresponds to a departure of less than $2.0$ standard deviations from the mean. Predictions from StackedGP are well inside the credible interval with a maximum average ratio of $0.41$. Here, $\hat{y}$ is the predicted mean, $y^*$ is the actual true value, and $\sigma$ is the analytical predicted standard deviation.

\subsection{Model Composition -  Forest Fire Dataset}
\begin{figure}[htp!]
	\centering
	\includegraphics[scale=0.4]{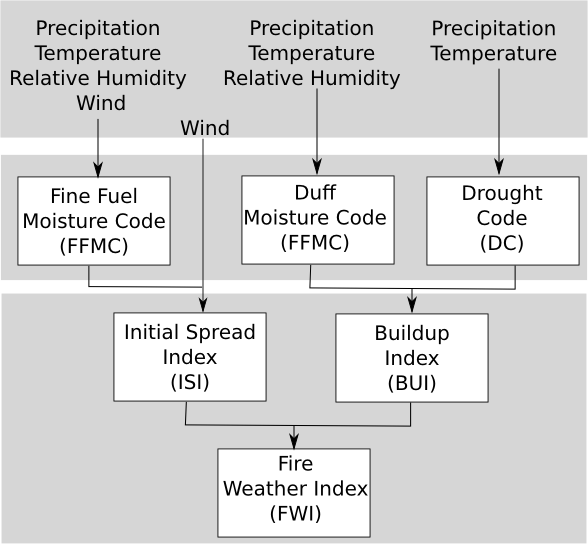}
	\caption{Example 2 (forest fire) - Structure of the fire weather index (FWI) system module of the Canadian forest fire danger rating system (\citet{taylor2006science}).}
	\label{fig:fwi} 
\end{figure}

The prediction of the burned area from forest fires has been discussed in different studies such as \citet{cortez2007data} and \citet{castelli2015}. The burned area of forest fires has been predicted using meteorological conditions (e.g. temperature, wind) and/or several Canadian forest fire weather indices (\citet{taylor2006science}) for rating fire danger, namely fine fuel moisture code (FFMC), duff moisture code (DMC), drought code (DC), initial spread index (ISI), and buildup index (BUI), as shown in Figure~\ref{fig:fwi}. 

\begin{figure}[htp!]
	\centering
	\input{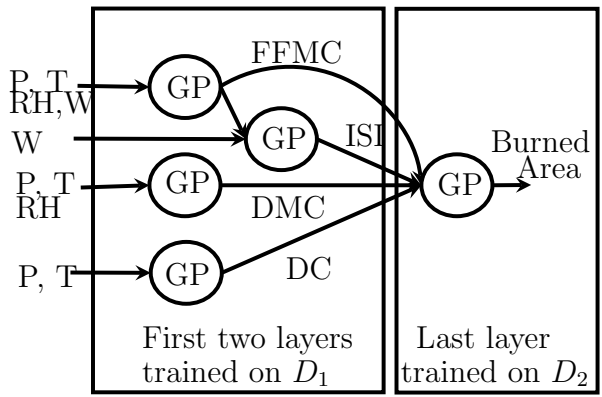}
	\caption{Example 2 (forest fire) - StackedGP for predicting burned area based on estimated FWI indices. Letters P, T, RH, W stands for precipitation, temperature, relative humidity and wind respectively. Also, the first two layers are trained using dataset $D_1$ , while dataset $D_2$ is used to train the last layer.}
	\label{fig:sgp_forest} 
\end{figure}

In this application we are interested in developing a StackedGP by first modeling the fire indices using meteorological variables $T$ from one dataset presented in \citet{van1985equations} and then model the burned area based on fire indices using another dataset presented in \citet{cortez2007data}. The proposed StackedGP is depicted in Figure~\ref{fig:sgp_forest}. The GP nodes corresponding to the four fire indices (FFMC,DMC, DC, and ISI) are trained from data published in \citet{van1985equations} according to the hierarchical structure shown in Figure~\ref{fig:fwi}. While the second dataset (\citet{cortez2007data}) contains meteorological conditions along with the fire indices and burned area, we assume that the meteorological conditions are missing in the training phase from this dataset and use only the fire indices and burned area data to train the GP node in the last layer of the StackedGP. 

A 10-fold cross validation is applied to the dataset published by \citet{cortez2007data} to train the burned area node and test the whole StackedGP model. Because of the skewed distribution of the burned area values and to ensure positive value for our predictions, instead of directly modeling the burned area using StackedGP, we have modeled the log of the burned area. As a result, the final mean and variance of the burned area $B[T]$ as a function of the meteorological conditions $T$ is given by Eqs.~\ref{log-normal:mean} and~\ref{log-normal:var} respectively. In additions, we have found that scaling the target variable to have zero mean and unit variance to be a beneficial preprocessing step.

\begin{equation}
	\label{log-normal:mean}
	E[B]=[e^{\sigma_{\ln B}^2}-1]e^{2\mu_{\ln B}+\sigma_{\ln B}^2}
\end{equation}
\begin{equation}
	\label{log-normal:var}
	Var[B]=e^{\mu_{\ln B}+0.5\sigma_{\ln B}^2}
\end{equation}
Here, $\mu_{\ln B}$ and $\sigma_{\ln B}$ are the output of the probabilistic analytical StackedGP (Eqs.~\ref{equ:generalPredictedmean} and \ref{equ:generalPredictedVar}) in the case of the RBF kernel, see Section~\ref{sec:RBFkernel}.

The result of modeling the burned area using the StackedGP is shown in Table~\ref{table:forestResults}. The StackedGP model is compared with the results of $5$ other regression models reported by \citet{cortez2007data}. Because these regression models have been tested using different input spaces, Table~\ref{table:forestResults} tabulates the best results achieved by each model as described in \citet{cortez2007data}. Even though the StackedGP predicts the burned area based on estimated indices from the first dataset and not the actual values as presented in the second dataset, it is still able to give comparable results with the other models that make use of meteorological conditions and/or fire indices available in the second dataset. This experiment emphasizes that the StackedGP is able to combine knowledge from different datasets with noticeable performance.
\begin{table}[h]
	\begin{center}
		\begin{tabular}{llll}
			{\bf Model}   			&{\bf Input} &{\bf MAE}	 	&{\bf RMSE}\\ \hline \\
			{\bf StackedGP}			&T			 &{12.80}		&{\bf 46.0}\\
			{ MR}					&FWI		 &{13}			&{64.5}\\			
			{ DT}					&T			 &{13.18}		&{64.5}\\	
			{ RF}					&T			 &{12.98}		&{64.4}\\
			{ NN}					&T			 &{13.08}		&{64.6}\\
			{ SVM}					&T			 &{\bf 12.71}	&{64.7}\\				    
		\end{tabular}		
	\end{center}
	\caption{Example 2 (forest fire) - Predictive results using different models. The input for each model is $T$ for meteorological features and $FWI$ for fire indices. Multiple regression (MR), decision trees (DT), random forests (RF), and neural networks (NN). } 
	\label{table:forestResults}
\end{table}

\subsection{Cascading Predictions - Jura Dataset}
In this subsection we use Jura dataset collected by the Swiss Federal Institute of Technology at Lauasanne (\citet{atteia1994,webster1994}). The dataset contains concentration samples of several heavy metals at $359$ different locations. Similar to previous experiments (\citet{goovaerts1997,alvarez2011,wilson12icml}), we are interested in predicting cadmium concentrations, the primary response at $100$ locations given $259$ training measurement points. The training data contains location information and concentrations of various metals (Cd, Zn, Ni, Cr, Co, Pb and Cu) at the sampled sites. The primary response is the concentration of Cd, and the other metals are considered secondary responses.

Note that standard Gaussian processes model each response variable independently and thus knowledge of secondary responses cannot help in predicting the primary one (\citet{seeger2005}). In this case a standard Gaussian process (StandardGP) will use a training dataset with only locations as inputs and Cd measurements as target (\citet{alvarez2011,wilson12icml}). Multi-output regression models such as co-kriging (\citet{cressie1992statistics}) can use the correlation between secondary and primary response to improve the prediction of Cd. The StackedGP, while it does not model the correlation between primary and secondary responses, it can be used to enhance the prediction of the primary response using intermediate predictions of the secondary responses.

In the heterotopic case (\citet{goovaerts1997}), the primary target is undersampled relative to the secondary variables. This provides access to  secondary information such as Ni and Zn at $100$ locations being estimated. As a result a standard Gaussian process can be built to have Ni and Zn directly as inputs. Here we will denote it as StandardGP(Zn,Ni). This is also the case for comparing our results with other six multi-task regression models as reported by \citet{wilson12icml} and tabulated in Table~\ref{table:jura_results}. 

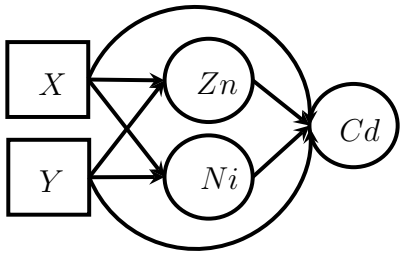
\begin{figure}[htp!]
	\centering
\ifx\du\undefined
  \newlength{\du}
\fi
\setlength{\du}{15\unitlength}
\begin{tikzpicture}[scale=0.6]
\pgftransformxscale{1.000000}
\pgftransformyscale{-1.000000}
\definecolor{dialinecolor}{rgb}{0.000000, 0.000000, 0.000000}
\pgfsetstrokecolor{dialinecolor}
\definecolor{dialinecolor}{rgb}{1.000000, 1.000000, 1.000000}
\pgfsetfillcolor{dialinecolor}
\definecolor{dialinecolor}{rgb}{1.000000, 1.000000, 1.000000}
\pgfsetfillcolor{dialinecolor}
\pgfpathellipse{\pgfpoint{15.962464\du}{13.211682\du}}{\pgfpoint{1.853364\du}{0\du}}{\pgfpoint{0\du}{1.751682\du}}
\pgfusepath{fill}
\pgfsetlinewidth{0.100000\du}
\pgfsetdash{}{0pt}
\pgfsetdash{}{0pt}
\pgfsetmiterjoin
\definecolor{dialinecolor}{rgb}{0.000000, 0.000000, 0.000000}
\pgfsetstrokecolor{dialinecolor}
\pgfpathellipse{\pgfpoint{15.962464\du}{13.211682\du}}{\pgfpoint{1.853364\du}{0\du}}{\pgfpoint{0\du}{1.751682\du}}
\pgfusepath{stroke}
\definecolor{dialinecolor}{rgb}{0.000000, 0.000000, 0.000000}
\pgfsetstrokecolor{dialinecolor}
\node at (15.962464\du,13.406682\du){};
\definecolor{dialinecolor}{rgb}{1.000000, 1.000000, 1.000000}
\pgfsetfillcolor{dialinecolor}
\pgfpathellipse{\pgfpoint{15.959264\du}{17.275082\du}}{\pgfpoint{1.853364\du}{0\du}}{\pgfpoint{0\du}{1.751682\du}}
\pgfusepath{fill}
\pgfsetlinewidth{0.100000\du}
\pgfsetdash{}{0pt}
\pgfsetdash{}{0pt}
\pgfsetmiterjoin
\definecolor{dialinecolor}{rgb}{0.000000, 0.000000, 0.000000}
\pgfsetstrokecolor{dialinecolor}
\pgfpathellipse{\pgfpoint{15.959264\du}{17.275082\du}}{\pgfpoint{1.853364\du}{0\du}}{\pgfpoint{0\du}{1.751682\du}}
\pgfusepath{stroke}
\definecolor{dialinecolor}{rgb}{0.000000, 0.000000, 0.000000}
\pgfsetstrokecolor{dialinecolor}
\node at (15.959264\du,17.470082\du){};
\pgfsetlinewidth{0.100000\du}
\pgfsetdash{}{0pt}
\pgfsetdash{}{0pt}
\pgfsetbuttcap
{
\definecolor{dialinecolor}{rgb}{0.000000, 0.000000, 0.000000}
\pgfsetfillcolor{dialinecolor}
\pgfsetarrowsend{stealth}
\definecolor{dialinecolor}{rgb}{0.000000, 0.000000, 0.000000}
\pgfsetstrokecolor{dialinecolor}
\draw (10.900000\du,13.175000\du)--(14.109100\du,13.211700\du);
}
\pgfsetlinewidth{0.100000\du}
\pgfsetdash{}{0pt}
\pgfsetdash{}{0pt}
\pgfsetbuttcap
{
\definecolor{dialinecolor}{rgb}{0.000000, 0.000000, 0.000000}
\pgfsetfillcolor{dialinecolor}
\pgfsetarrowsend{stealth}
\definecolor{dialinecolor}{rgb}{0.000000, 0.000000, 0.000000}
\pgfsetstrokecolor{dialinecolor}
\draw (10.900000\du,13.175000\du)--(14.105900\du,17.275100\du);
}
\pgfsetlinewidth{0.100000\du}
\pgfsetdash{}{0pt}
\pgfsetdash{}{0pt}
\pgfsetbuttcap
{
\definecolor{dialinecolor}{rgb}{0.000000, 0.000000, 0.000000}
\pgfsetfillcolor{dialinecolor}
\pgfsetarrowsend{stealth}
\definecolor{dialinecolor}{rgb}{0.000000, 0.000000, 0.000000}
\pgfsetstrokecolor{dialinecolor}
\draw (10.940000\du,17.285000\du)--(14.105900\du,17.275100\du);
}
\definecolor{dialinecolor}{rgb}{1.000000, 1.000000, 1.000000}
\pgfsetfillcolor{dialinecolor}
\pgfpathellipse{\pgfpoint{22.051164\du}{15.111682\du}}{\pgfpoint{1.853364\du}{0\du}}{\pgfpoint{0\du}{1.751682\du}}
\pgfusepath{fill}
\pgfsetlinewidth{0.100000\du}
\pgfsetdash{}{0pt}
\pgfsetdash{}{0pt}
\pgfsetmiterjoin
\definecolor{dialinecolor}{rgb}{0.000000, 0.000000, 0.000000}
\pgfsetstrokecolor{dialinecolor}
\pgfpathellipse{\pgfpoint{22.051164\du}{15.111682\du}}{\pgfpoint{1.853364\du}{0\du}}{\pgfpoint{0\du}{1.751682\du}}
\pgfusepath{stroke}
\definecolor{dialinecolor}{rgb}{0.000000, 0.000000, 0.000000}
\pgfsetstrokecolor{dialinecolor}
\node at (22.051164\du,15.306682\du){};
\definecolor{dialinecolor}{rgb}{0.000000, 0.000000, 0.000000}
\pgfsetstrokecolor{dialinecolor}
\node[anchor=west] at (9.193360\du,17.261700\du){};
\definecolor{dialinecolor}{rgb}{0.000000, 0.000000, 0.000000}
\pgfsetstrokecolor{dialinecolor}
\node[anchor=west] at (14.915000\du,13.330000\du){$Zn$};
\definecolor{dialinecolor}{rgb}{0.000000, 0.000000, 0.000000}
\pgfsetstrokecolor{dialinecolor}
\node[anchor=west] at (15.080000\du,17.340000\du){$Ni$};
\definecolor{dialinecolor}{rgb}{0.000000, 0.000000, 0.000000}
\pgfsetstrokecolor{dialinecolor}
\node[anchor=west] at (20.940000\du,15.305000\du){$Cd$};
\pgfsetlinewidth{0.100000\du}
\pgfsetdash{}{0pt}
\pgfsetdash{}{0pt}
\pgfsetmiterjoin
\definecolor{dialinecolor}{rgb}{1.000000, 1.000000, 1.000000}
\pgfsetfillcolor{dialinecolor}
\fill (7.500000\du,11.600000\du)--(7.500000\du,14.750000\du)--(10.900000\du,14.750000\du)--(10.900000\du,11.600000\du)--cycle;
\definecolor{dialinecolor}{rgb}{0.000000, 0.000000, 0.000000}
\pgfsetstrokecolor{dialinecolor}
\draw (7.500000\du,11.600000\du)--(7.500000\du,14.750000\du)--(10.900000\du,14.750000\du)--(10.900000\du,11.600000\du)--cycle;
\definecolor{dialinecolor}{rgb}{0.000000, 0.000000, 0.000000}
\pgfsetstrokecolor{dialinecolor}
\node[anchor=west] at (8.290000\du,13.355000\du){$X$};
\pgfsetlinewidth{0.100000\du}
\pgfsetdash{}{0pt}
\pgfsetdash{}{0pt}
\pgfsetmiterjoin
\definecolor{dialinecolor}{rgb}{1.000000, 1.000000, 1.000000}
\pgfsetfillcolor{dialinecolor}
\fill (7.540000\du,15.710000\du)--(7.540000\du,18.860000\du)--(10.940000\du,18.860000\du)--(10.940000\du,15.710000\du)--cycle;
\definecolor{dialinecolor}{rgb}{0.000000, 0.000000, 0.000000}
\pgfsetstrokecolor{dialinecolor}
\draw (7.540000\du,15.710000\du)--(7.540000\du,18.860000\du)--(10.940000\du,18.860000\du)--(10.940000\du,15.710000\du)--cycle;
\definecolor{dialinecolor}{rgb}{0.000000, 0.000000, 0.000000}
\pgfsetstrokecolor{dialinecolor}
\node[anchor=west] at (8.315000\du,17.430000\du){$Y$};
\pgfsetlinewidth{0.100000\du}
\pgfsetdash{}{0pt}
\pgfsetdash{}{0pt}
\pgfsetbuttcap
{
\definecolor{dialinecolor}{rgb}{0.000000, 0.000000, 0.000000}
\pgfsetfillcolor{dialinecolor}
\pgfsetarrowsend{stealth}
\definecolor{dialinecolor}{rgb}{0.000000, 0.000000, 0.000000}
\pgfsetstrokecolor{dialinecolor}
\draw (17.812600\du,17.275100\du)--(20.197800\du,15.111700\du);
}
\pgfsetlinewidth{0.100000\du}
\pgfsetdash{}{0pt}
\pgfsetdash{}{0pt}
\pgfsetbuttcap
{
\definecolor{dialinecolor}{rgb}{0.000000, 0.000000, 0.000000}
\pgfsetfillcolor{dialinecolor}
\pgfsetarrowsend{stealth}
\definecolor{dialinecolor}{rgb}{0.000000, 0.000000, 0.000000}
\pgfsetstrokecolor{dialinecolor}
\draw (17.815800\du,13.211700\du)--(20.197800\du,15.111700\du);
}
\pgfsetlinewidth{0.100000\du}
\pgfsetdash{}{0pt}
\pgfsetdash{}{0pt}
\pgfsetbuttcap
{
\definecolor{dialinecolor}{rgb}{0.000000, 0.000000, 0.000000}
\pgfsetfillcolor{dialinecolor}
\pgfsetarrowsend{stealth}
\definecolor{dialinecolor}{rgb}{0.000000, 0.000000, 0.000000}
\pgfsetstrokecolor{dialinecolor}
\draw (10.940000\du,17.285000\du)--(14.109100\du,13.211700\du);
}
\pgfsetlinewidth{0.100000\du}
\pgfsetdash{}{0pt}
\pgfsetdash{}{0pt}
\pgfsetbuttcap
{
\definecolor{dialinecolor}{rgb}{0.000000, 0.000000, 0.000000}
\pgfsetfillcolor{dialinecolor}
\pgfsetarrowsstart{stealth}
\definecolor{dialinecolor}{rgb}{0.000000, 0.000000, 0.000000}
\pgfsetstrokecolor{dialinecolor}
\pgfpathmoveto{\pgfpoint{20.197784\du}{15.112205\du}}
\pgfpatharc{2}{-158}{4.821120\du and 4.821120\du}
\pgfusepath{stroke}
}
\pgfsetlinewidth{0.100000\du}
\pgfsetdash{}{0pt}
\pgfsetdash{}{0pt}
\pgfsetbuttcap
{
\definecolor{dialinecolor}{rgb}{0.000000, 0.000000, 0.000000}
\pgfsetfillcolor{dialinecolor}
\pgfsetarrowsend{stealth}
\definecolor{dialinecolor}{rgb}{0.000000, 0.000000, 0.000000}
\pgfsetstrokecolor{dialinecolor}
\pgfpathmoveto{\pgfpoint{10.939889\du}{17.284736\du}}
\pgfpatharc{158}{-3}{4.822244\du and 4.822244\du}
\pgfusepath{stroke}
}
\end{tikzpicture}
	\caption{Example 3 (cascading predictions) - StackedGP for predicting $Cd$ based on estimated $Zn$ and $Ni$ at location of interest $X$ and $Y$.}
	\label{fig:sgp_cd} 
\end{figure}

The first proposed StackedGP uses the first layer to model Zn and Ni based on  locations and the second layer to model Cd based on the locations and the estimated output of the first layer, see Figure~\ref{fig:sgp_cd}. In the heterotopic case the StackedGP can use directly the available measurements of Ni and Zn instead of predictions by setting the uncertainty associated with these measurements to zero. In this case the StackedGP acts as the StandardGP(Zn,Ni).

\begin{figure}[htp!]
	\centering
	\input{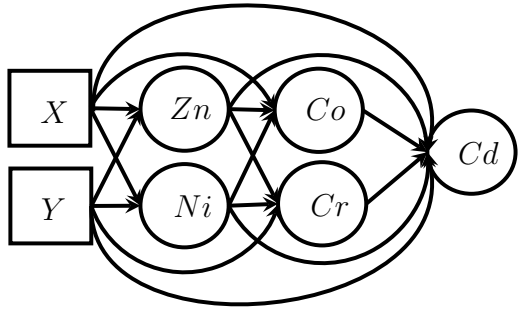}
	\caption{Example 3 (cascading predictions) - StackedGP for predicting $Cd$ based on estimated $Zn$, $Ni$, $Co$, and $Cr$ at location of interest $X$ and $Y$.}
	\label{fig:sgp_co_cr_cd} 
\end{figure}

Three other structures are proposed by using intermediate predictions of Co, Cr, and Co and Cr together. In this case, we have a three layer StackedGP to model Cd, see Figure~\ref{fig:sgp_co_cr_cd}. The first layer is the same as in the previous setup. The second layer models intermediate responses (Co, Cr, and Co and Cr). The third layer is used to model Cd based on the second layer predictions in additions to the input/output of the first layer, namely location and Zn and Ni. 

Table~\ref{table:jura_results} shows the results of these stacked structures, StackedGP(Co), StackedGP(Cr) and StackedGP(Co,Cr). While measurements of Ni and Zn are available in the testing scenarios, there are no measurements for Co and Cr during testing. Thus, Cd predictions of these three StackedGPs rely on predictions of Co and Cr using locations and Ni and Zn measurements at these locations.

\begin{table}[h]
	\begin{center}
		\begin{tabular}{ll}
			{\bf Model}   				&{\bf MAE} \\ \hline \\
			{\bf StackedGP} 			&{0.3833}\\
			{\bf StackedGP(Co)} 		&{0.3617}\\  
			{\bf StackedGP(Cr)}  		&{0.3884}\\  
			{\bf StackedGP(Co,Cr)}  	&{\textbf{0.3602}}\\ 
			GPRN(VB)~\citet{wilson12icml}&0.4040 \\
			SLFM(VB)~\citet{seeger2005}  &0.4247 \\
			SLFM~\citet{seeger2005}      &0.4578 \\
			ICM~\citet{goovaerts1997}	&0.4608 \\
			CMOGP~\citet{alvarez2011}		&0.4552 \\
			Co-Kriging	 				&0.51 \\
			StandardGP(Zn,Ni) 			&{0.3833}\\
			StandardGP 					&{0.5714}\\             
		\end{tabular}		
	\end{center}
	\caption{Example 3 (cascading predictions) - Performance on modeling Cd using different two/three layers StackedGP structures with mean absolute error (MAE) as performance metric.}
	\label{table:jura_results}
\end{table}

The mean absolute error (MAE) between the true and estimated Cd is calculated at the $100$ target locations. Overall StackedGP gives better results as compared with the other models. Also, when Zn and Ni measurements are available as assumed by the other multi-output regression models (\citet{wilson12icml,alvarez2011}), then a StandardGP(Ni,Zn) can provide a lower MAE than the other six multi-output regression models. However, StackedGP can provide a better performance over the Standard(Zn,Ni) by making use of intermediate predictions of secondary responses.

The complexity of most of multi-task models is cubic in both the number of output responses and size of the training dataset (\citet{wilson12icml}). However, StackedGP scales linearly with the number of nodes in the structure because of the independent training of the nodes, which can be easily parallelized. In additions, sparse approximation techniques can be used to further reduce this complexity in the case of large training datasets (\citet{snelson2007,damianou2011variational}).

For all these experiments we found that the log transformation and normalization can lead to better results. For multi-responses in the middle layer, we used independent component analysis (ICA) to obtain independent projections of secondary responses. This is required as the current derivation assumes that inputs to a GP node are independent.

\subsection{Uncertainty Propagation - Atmospheric Transport}
To motivate the concept of uncertainty propagation in atmospheric transport, we consider a simple advection of a 2D Gaussian-shaped puff (\citet{Nielsen1999,TerejanuP_IF_2007}). The states of the puff evolve using the following equations.
\begin{align}
x_{k+1} = x_{k}+u_{x}(x_k)\Delta t \\
y_{k+1} = y_{k}+u_{y}(y_k)\Delta t \\
d_{k+1} = d_{k}+\sqrt{u^2_{x}(x_k)+u^2_{y}(y_k)}\Delta t
\end{align}

Here, $(x_{k}, y_{k})$ is the position of the center of the puff, and the downwind distance from the source $d_{k}$ is used to compute the puff radius, $\sigma_k=p d_k^q$ in models such as RIMPUFF (\citet{Nielsen1999}) based on Karlsruhe-J\"{u}lich diffusion coefficients (\citet{Reddy2007}), $(p,q)$.

The goal here is to build a GP emulator for the above dynamical system, knowing that the release location is fixed at ($x_0=0$km, $y_0=0$km) and the wind velocity is uncertain with normally distributed wind components ($u_x, u_y$).
\begin{equation}
u_x, u_y \sim \mathcal{N}(4\textrm{m/s}, 1\textrm{m/s})
\label{eq:wind_distribution}
\end{equation}

The GP emulator $h(\cdot)$ is constructed using $15$ training trajectories that start at the same release location, but correspond to different wind fields that randomly sampled from the distribution in Eq.~\ref{eq:wind_distribution}. The total simulation time is $30$min with a time step $\Delta t = 90$sec. 
\begin{equation}
[x_{k+1}, y_{k+1}, d_{k+1}] = h(x_k, y_k, u_x(x_k), u_y(x_k))
\end{equation}

Another GP model is constructed to determine the wind field based on $16$ wind sensors positioned $4$km apart in both directions. The wind sensor readings are just independent and identically distributed samples from Eq.~\ref{eq:wind_distribution}.
\begin{equation}
[u_x(x), u_y(y)] = g(x,y)
\end{equation}

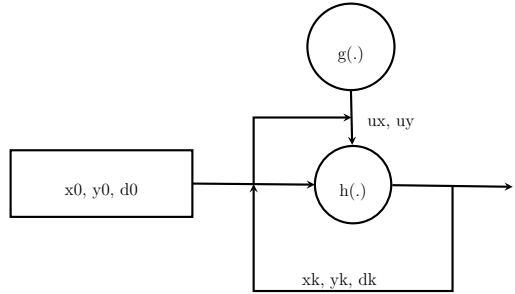
\begin{figure}[h]
	\centering
	\resizebox{.5\linewidth}{!}{
\ifx\du\undefined
  \newlength{\du}
\fi
\setlength{\du}{15\unitlength}
\begin{tikzpicture}
\pgftransformxscale{1.000000}
\pgftransformyscale{-1.000000}
\definecolor{dialinecolor}{rgb}{0.000000, 0.000000, 0.000000}
\pgfsetstrokecolor{dialinecolor}
\definecolor{dialinecolor}{rgb}{1.000000, 1.000000, 1.000000}
\pgfsetfillcolor{dialinecolor}
\definecolor{dialinecolor}{rgb}{1.000000, 1.000000, 1.000000}
\pgfsetfillcolor{dialinecolor}
\pgfpathellipse{\pgfpoint{23.425000\du}{12.550000\du}}{\pgfpoint{1.775000\du}{0\du}}{\pgfpoint{0\du}{1.800000\du}}
\pgfusepath{fill}
\pgfsetlinewidth{0.100000\du}
\pgfsetdash{}{0pt}
\pgfsetdash{}{0pt}
\pgfsetmiterjoin
\definecolor{dialinecolor}{rgb}{0.000000, 0.000000, 0.000000}
\pgfsetstrokecolor{dialinecolor}
\pgfpathellipse{\pgfpoint{23.425000\du}{12.550000\du}}{\pgfpoint{1.775000\du}{0\du}}{\pgfpoint{0\du}{1.800000\du}}
\pgfusepath{stroke}
\definecolor{dialinecolor}{rgb}{0.000000, 0.000000, 0.000000}
\pgfsetstrokecolor{dialinecolor}
\node at (23.425000\du,12.894444\du){h(.)};
\definecolor{dialinecolor}{rgb}{1.000000, 1.000000, 1.000000}
\pgfsetfillcolor{dialinecolor}
\pgfpathellipse{\pgfpoint{23.325000\du}{6.137745\du}}{\pgfpoint{2.025000\du}{0\du}}{\pgfpoint{0\du}{2.012255\du}}
\pgfusepath{fill}
\pgfsetlinewidth{0.100000\du}
\pgfsetdash{}{0pt}
\pgfsetdash{}{0pt}
\pgfsetmiterjoin
\definecolor{dialinecolor}{rgb}{0.000000, 0.000000, 0.000000}
\pgfsetstrokecolor{dialinecolor}
\pgfpathellipse{\pgfpoint{23.325000\du}{6.137745\du}}{\pgfpoint{2.025000\du}{0\du}}{\pgfpoint{0\du}{2.012255\du}}
\pgfusepath{stroke}
\definecolor{dialinecolor}{rgb}{0.000000, 0.000000, 0.000000}
\pgfsetstrokecolor{dialinecolor}
\node at (23.325000\du,6.482190\du){g(.)};
\pgfsetlinewidth{0.100000\du}
\pgfsetdash{}{0pt}
\pgfsetdash{}{0pt}
\pgfsetbuttcap
{
\definecolor{dialinecolor}{rgb}{0.000000, 0.000000, 0.000000}
\pgfsetfillcolor{dialinecolor}
\pgfsetarrowsend{stealth}
\definecolor{dialinecolor}{rgb}{0.000000, 0.000000, 0.000000}
\pgfsetstrokecolor{dialinecolor}
\draw (25.249527\du,12.574573\du)--(30.850000\du,12.650000\du);
}
\pgfsetlinewidth{0.100000\du}
\pgfsetdash{}{0pt}
\pgfsetdash{}{0pt}
\pgfsetbuttcap
{
\definecolor{dialinecolor}{rgb}{0.000000, 0.000000, 0.000000}
\pgfsetfillcolor{dialinecolor}
\pgfsetarrowsend{stealth}
\definecolor{dialinecolor}{rgb}{0.000000, 0.000000, 0.000000}
\pgfsetstrokecolor{dialinecolor}
\draw (23.325000\du,8.150000\du)--(23.382983\du,10.701270\du);
}
\pgfsetlinewidth{0.100000\du}
\pgfsetdash{}{0pt}
\pgfsetdash{}{0pt}
\pgfsetbuttcap
{
\definecolor{dialinecolor}{rgb}{0.000000, 0.000000, 0.000000}
\pgfsetfillcolor{dialinecolor}
\pgfsetarrowsend{stealth}
\definecolor{dialinecolor}{rgb}{0.000000, 0.000000, 0.000000}
\pgfsetstrokecolor{dialinecolor}
\draw (15.951250\du,12.500000\du)--(21.650000\du,12.550000\du);
}
\definecolor{dialinecolor}{rgb}{1.000000, 1.000000, 1.000000}
\pgfsetfillcolor{dialinecolor}
\fill (7.498750\du,10.950000\du)--(7.498750\du,14.050000\du)--(15.951250\du,14.050000\du)--(15.951250\du,10.950000\du)--cycle;
\pgfsetlinewidth{0.100000\du}
\pgfsetdash{}{0pt}
\pgfsetdash{}{0pt}
\pgfsetmiterjoin
\definecolor{dialinecolor}{rgb}{0.000000, 0.000000, 0.000000}
\pgfsetstrokecolor{dialinecolor}
\draw (7.498750\du,10.950000\du)--(7.498750\du,14.050000\du)--(15.951250\du,14.050000\du)--(15.951250\du,10.950000\du)--cycle;
\definecolor{dialinecolor}{rgb}{0.000000, 0.000000, 0.000000}
\pgfsetstrokecolor{dialinecolor}
\node at (11.725000\du,12.844444\du){x0, y0, d0};
\pgfsetlinewidth{0.100000\du}
\pgfsetdash{}{0pt}
\pgfsetdash{}{0pt}
\pgfsetmiterjoin
\pgfsetbuttcap
{
\definecolor{dialinecolor}{rgb}{0.000000, 0.000000, 0.000000}
\pgfsetfillcolor{dialinecolor}
\pgfsetarrowsend{stealth}
{\pgfsetcornersarced{\pgfpoint{0.000000\du}{0.000000\du}}\definecolor{dialinecolor}{rgb}{0.000000, 0.000000, 0.000000}
\pgfsetstrokecolor{dialinecolor}
\draw (18.800625\du,12.525000\du)--(18.800625\du,9.425635\du)--(23.353992\du,9.425635\du);
}}
\pgfsetlinewidth{0.100000\du}
\pgfsetdash{}{0pt}
\pgfsetdash{}{0pt}
\pgfsetmiterjoin
\pgfsetbuttcap
{
\definecolor{dialinecolor}{rgb}{0.000000, 0.000000, 0.000000}
\pgfsetfillcolor{dialinecolor}
\pgfsetarrowsend{stealth}
{\pgfsetcornersarced{\pgfpoint{0.000000\du}{0.000000\du}}\definecolor{dialinecolor}{rgb}{0.000000, 0.000000, 0.000000}
\pgfsetstrokecolor{dialinecolor}
\draw (28.049763\du,12.612286\du)--(28.049763\du,17.500000\du)--(18.800625\du,17.500000\du)--(18.800625\du,12.525000\du);
}}
\definecolor{dialinecolor}{rgb}{0.000000, 0.000000, 0.000000}
\pgfsetstrokecolor{dialinecolor}
\node[anchor=west] at (20.750000\du,17.000000\du){xk, yk, dk};
\definecolor{dialinecolor}{rgb}{0.000000, 0.000000, 0.000000}
\pgfsetstrokecolor{dialinecolor}
\node[anchor=west] at (23.790000\du,9.725000\du){ux, uy};
\end{tikzpicture}}
	\caption{Example 4 (uncertainty propagation) - StackedGP model for uncertainty propagation using emulated 2D puff advection driven by uncertain wind field.}
	\label{fig:sgp_puff}
\end{figure}

Note, that in this particular case the wind velocity at different locations is correlated. Both emulators use RBF kernels, and they are stacked to build a recurrent StackedGP as shown in Figure~\ref{fig:sgp_puff}.

To assess the effect of the two assumptions in constructing the StackedGP (independent inputs for each layer and Gaussian distribution approximation for the output of each layer), we compare the approximate mean and variance of the puff states from StackedGP using the proposed algorithm with those resulted from a Monte Carlo propagation of uncertainty through the StackedGP using $1000$ samples.

\footnotesize
\begin{table}[H]
	\footnotesize
	\begin{tabular}{ |l|l|l|l|l|l|l|l|l|l|l|l|l| }
		\hline
		& \multicolumn{6}{|c|}{Approximate Propagation} & \multicolumn{6}{|c|}{Monte Carlo} \\ \hline 
		$k$ & \multicolumn{2}{|c|}{$x_k$} & \multicolumn{2}{|c|}{$y_k$} & \multicolumn{2}{|c|}{$d_k$}& \multicolumn{2}{|c|}{$x_k$} & \multicolumn{2}{|c|}{$y_k$} & \multicolumn{2}{|c|}{$d_k$} \\ \hline
		& $\mu$ & $\sigma$ & $\mu$ & $\sigma$ & $\mu$ & $\sigma$ & $\mu$ & $\sigma$ & $\mu$ & $\sigma$ & $\mu$ & $\sigma$ \\ \hline
		$5$& $7.76$ & $0.19$& $7.83$ & $0.13$& $2.61$ & $0.23$& $7.74$ & $0.19$& $7.89$ & $0.14$& $2.59$ & $0.16$\\ \hline
		$10$& $9.52$ & $0.26$& $9.67$ & $0.18$& $5.19$ & $0.29$& $9.47$ & $0.27$& $9.76$ & $0.2$& $5.17$ & $0.24$\\ \hline
		$15$& $11.3$ & $0.32$& $11.5$ & $0.22$& $7.74$ & $0.34$& $11.22$ & $0.33$& $11.59$ & $0.24$& $7.72$ & $0.29$\\ \hline
		$20$& $13.09$ & $0.37$& $13.33$ & $0.26$& $10.26$ & $0.39$& $12.95$ & $0.38$& $13.38$ & $0.28$& $10.25$ & $0.33$\\ \hline
	\end{tabular}
	\caption{Example 4 (uncertainty propagation) - Predicted mean and standard deviation of puff states using proposed approximate and Monte Carlo propagation of uncertainty through StackedGP.}
	\label{table:atmospheric_dispersion}
\end{table}
\normalsize

	\begin{figure}[H]
		\begin{subfigure}{0.35\textwidth}
			\includegraphics[scale=0.23]{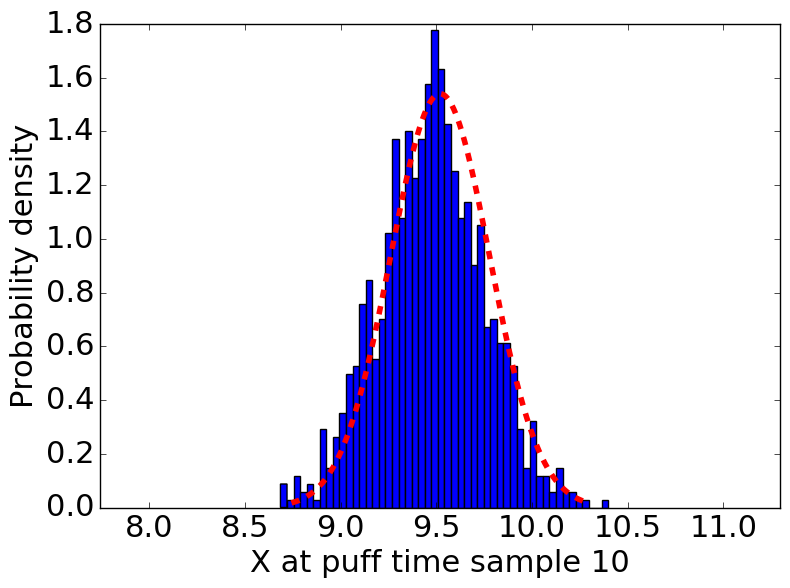}
			\caption{X location}
		\end{subfigure}
		\begin{subfigure}{0.33\textwidth}
			\includegraphics[scale=0.23]{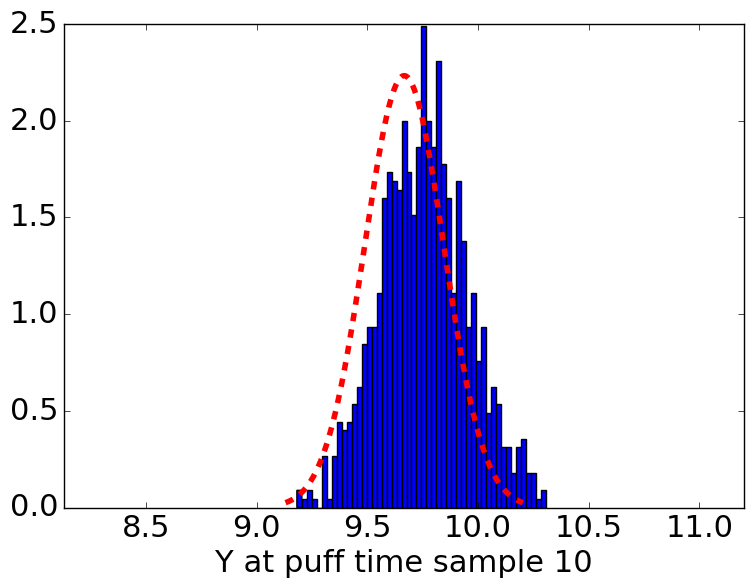}
			\caption{Y location}
		\end{subfigure}
		\begin{subfigure}{0.30\textwidth}
			\includegraphics[scale=0.23]{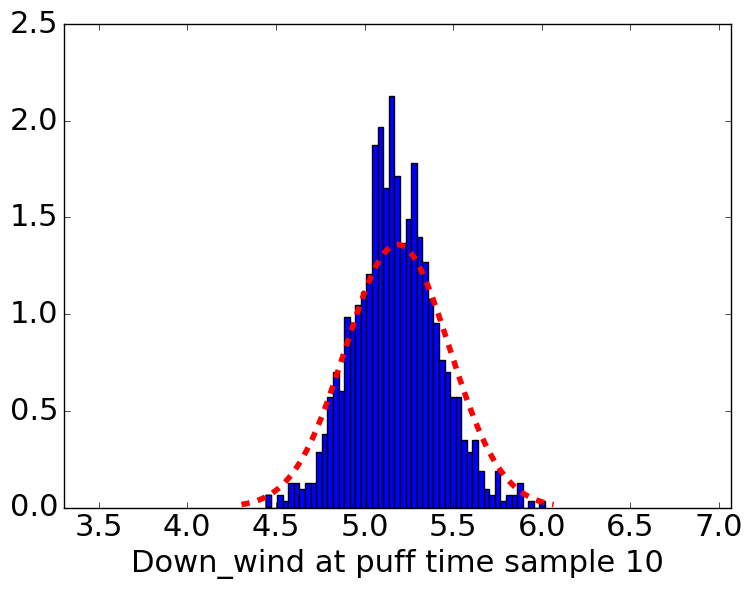}
			\caption{Down wind}
		\end{subfigure}
		\caption{Example 4 (uncertainty propagation) - Histogram of $1000$ MC samples (blue) and the predicted StackedGP Gaussian distribution (red) at time step $10$}
		\label{fig:compare_10}
	\end{figure}
	\begin{figure}[H]
		\begin{subfigure}{0.35\textwidth}
			\includegraphics[scale=0.23]{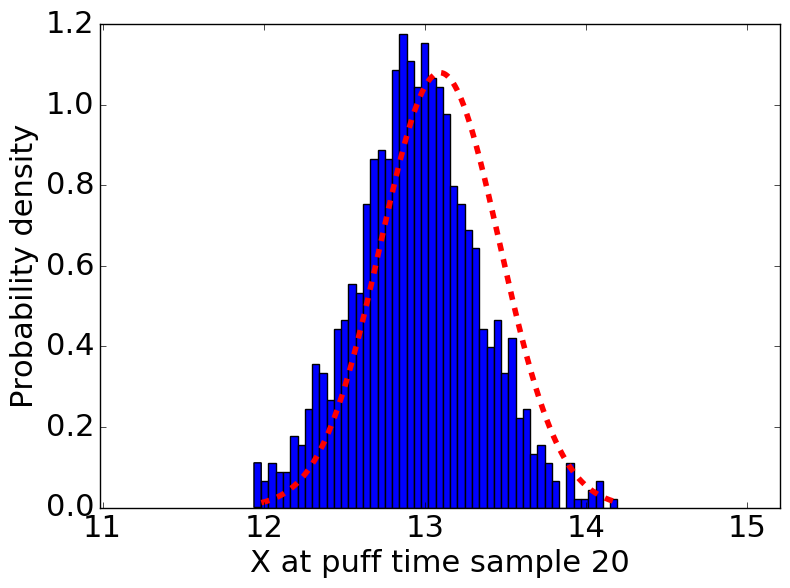}
			\caption{X location}
		\end{subfigure}
		\begin{subfigure}{0.33\textwidth}
			\includegraphics[scale=0.23]{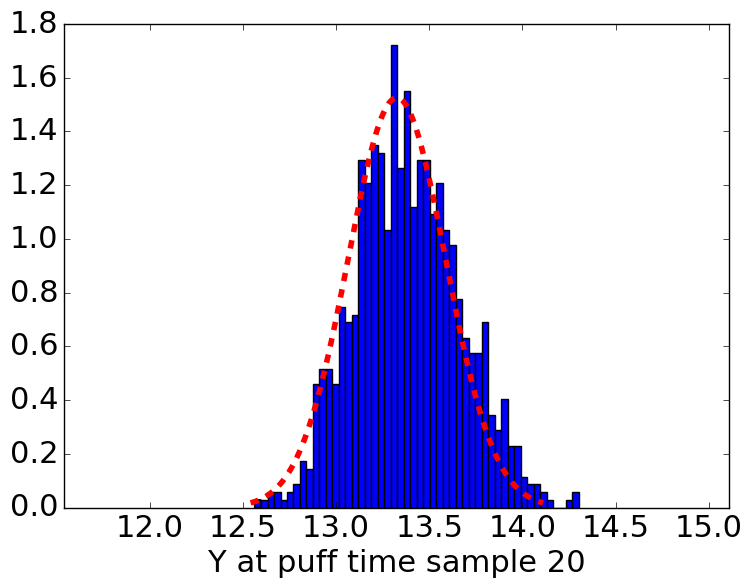}
			\caption{Y location}
		\end{subfigure}
		\begin{subfigure}{0.30\textwidth}
			\includegraphics[scale=0.23]{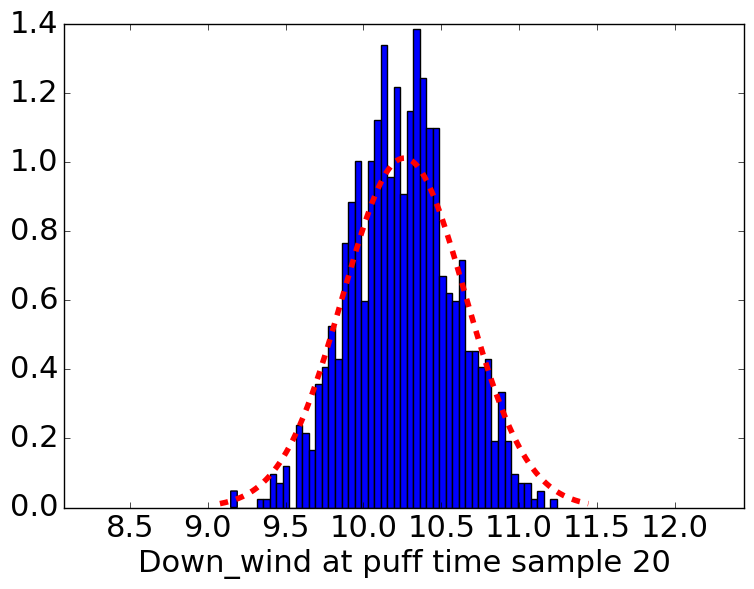}
			\caption{Down wind}
		\end{subfigure}
		\caption{Example 4 (uncertainty propagation) - Histogram of $1000$ MC samples (blue) and the predicted StackedGP Gaussian distribution (red) at time step $20$}
		\label{fig:compare_20}
	\end{figure}

Figures~\ref{fig:compare_10} and \ref{fig:compare_20} show the approximate predicted Gaussian distribution of the states along with the histogram of the Monte Carlo samples propagated through the StackedGP. Table~\ref{table:atmospheric_dispersion} lists the predicted mean and standard deviation of the puff states at different time steps.

Note that even though the state equations for the location of the puff are linear, because they are emulated using a GP, which at its turn is driven by a GP model for the wind field, the distribution of the StackedGP output may depart from the Gaussian distribution. The assumption of approximating the output with a Gaussian distribution may result in biasing the mean location. The statistical significant difference between the StackedGP approximate mean propagation and its Monte Carlo estimate confirms the impact of this approximation as shown in Table~\ref{table:atmospheric_dispersion}. 

Furthermore, the assumption of ignoring the correlation structure between the outputs of StackedGP may result in an artificial inflation of the uncertainty. In our simple example, this is clearly manifested in larger standard deviations for the downwind using approximate propagation as compared with the Monte Carlo estimate. This impact on uncertainty propagation might be exacerbated when more nonlinear models are used, which limits the horizon of uncertainty propagation. Obviously, the gain in computational speed combined with field measurements in the context of data assimilation may position these stacked model as real contenders for real time applications. We plan to investigate in the future the application of StackedGP to data assimilation. 

\section{Conclusions}
\label{sec:conclusions}
A stacked model of independently trained Gaussian processes, called StackedGP, is proposed as a modeling framework in the context of model composition. This is especially of interest in environmental modeling where, e.g., model composition is used to generate large scale predictions by combining geographical interpolation models with phenomenological models developed in the lab. An approximate approach is developed to obtain estimates of the quantities of interest with quantified uncertainties. This leverages the analytical moments of a Gaussian process with uncertain inputs when squared exponential and polynomial kernels are used. The StackedGP can be extended to any number of nodes and layers and has no restriction in selecting a suitable kernel for the input nodes. 

The numerical results show the utility of using StackedGP to learn from different datasets and propagate the uncertainty to quantities of interest. While it is not specifically designed to model correlations between secondary and primary responses, StackedGP can be used to enhance the prediction of primary responses by creating an intermediate layer of predictions of secondary responses. This comes with a lower computational complexity as compared with multi-output methods - and can make use of off-the-shelves Gaussian processes. While in the current paper we assume that outputs of intermediate layers are independent and resolve this using independent component analysis preprocessing, we plan to extend our derivation to account for these correlations in the next study. This will allow multi-output models to act as nodes in the proposed StackedGP.  

\section*{Acknowledgments}
This material is based upon work supported by the National Science Foundation under Grand No. 1504728 and 1632824. Dr. Terejanu has been supported by the National Institute of Food and Agriculture (NIFA)/USDA under Grand No. 2017-67017-26167.

\section*{Reference}
\bibliographystyle{elsarticle-harv}
\bibliography{refer}






%
%
%
\end{document}